\documentclass[letterpaper, 10 pt, conference]{ieeeconf}  

\IEEEoverridecommandlockouts                              

\usepackage{amsmath,amsfonts}
\usepackage{algorithmic}
\usepackage{algorithm}
\usepackage{array}
\usepackage{textcomp}
\usepackage{stfloats}
\usepackage{url}
\usepackage{verbatim}
\usepackage{graphicx}
\usepackage{cite}
\usepackage[per-mode=symbol]{siunitx}
\usepackage{gensymb}
\usepackage{xcolor}
\usepackage{tabularray}
\usepackage{graphics}
\usepackage{booktabs}
\usepackage{multirow}
\usepackage{threeparttable}
\usepackage{overpic}
\usepackage{makecell}
\usepackage{hyperref}
\usepackage{microtype}

\title{\LARGE \bf
LDHP: Library-Driven Hierarchical Planning for Non-prehensile Dexterous Manipulation
}

\author{Tierui He, Jiahui Zuo, Fumin Zhang, and Chao Zhao
\thanks{C. Zhao is with Jilin University, Changchun, China. T. He is with the Shanghai Marine Equipment Research Institute, Shanghai 200031, China. J. Zuo and F. Zhang are with The Hong Kong University of Science and Technology, Clear Water Bay, Hong Kong. Corresponding author: Chao Zhao (e-mail: chaozhao@jlu.edu.cn).}
}

\begin{document}

\maketitle
\thispagestyle{empty}
\pagestyle{empty}

\begin{abstract}

Non-prehensile manipulation is essential for handling thin, large, or otherwise ungraspable objects in unstructured settings. Prior planning and search-based methods often rely on ad-hoc manual designs or generate physically unrealizable motions by ignoring critical gripper properties, while training-based approaches are data-intensive and struggle to generalize to novel, out-of-distribution tasks. We propose a library-driven hierarchical planner (LDHP) that makes executability a first-class design goal: a top-tier contact-state planner proposes object-pose paths using \emph{MoveObject} primitives, and a bottom-tier grasp planner synthesizes feasible grasp sequences with \emph{AdjustGrasp} primitives; feasibility is certified by collision checks and quasi-static mechanics, and contact-sensitive segments are recovered via a bounded dichotomy refinement. This gripper-aware decomposition decouples object motion from grasp realizability, yields a task-agnostic pipeline that transfers across manipulation tasks and geometric variations without re-design, and exposes clean hooks for optional learned priors. Real-robot studies on zero-mobility lifting and slot insertion demonstrate consistent execution and robustness to shape and environment changes. 

\end{abstract}

\section{INTRODUCTION}

\begin{figure}[!t]
    \centering
    \includegraphics[width=1\linewidth]{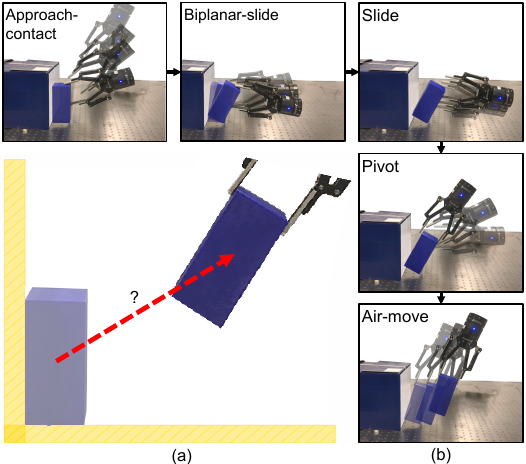}
    \caption{
    (a) A representative non-prehensile scenario where a block must be lifted from the table edge with no finger clearance;
    (b) LDHP synthesize an executable plan; example snapshots illustrate the primitive sequence.}
    \label{fg1}
\vspace{-20pt}
\end{figure}

Non-prehensile manipulation—pushing, sliding, toppling, and flipping—extends robotic capability well beyond pick-and-place and is central to long-horizon, contact-rich tasks (Fig. \ref{fg1}). Over the years, two lines of research have advanced the field in complementary ways. Planning-based approaches have made solid progress by explicitly modeling geometry and contact: task-and-motion planning with contact-state reasoning, quasi-static and dynamic analyses under Coulomb friction, and carefully engineered action primitives have enabled reproducible pipelines with physical guarantees \cite{m2}. In parallel, learning-based approaches reduce manual modeling effort and can discover contact strategies that are hard to hand-design by absorbing high-dimensional perception into compact policies \cite{l1,l2,l3}. 

Yet, practical deployment of manipulation pipelines reveals several persistent obstacles across existing paradigms. First, training-based approaches necessitate a substantial volume of data and considerable time investment. Moreover, they struggle to generalize to novel, unseen tasks that lie far outside the training distribution; as a representative example, it is particularly challenging to grasp an undecagonal object using a two-fingered parallel gripper that has lost its opening and closing actuation capability. Second, for manipulation planning approaches other than training-based ones—especially for non-prehensile tasks—ad-hoc methods are commonly adopted, where specific strategies must be manually predefined for particular tasks \cite{kim2019shallow, he2021scooping}. Third, searching-based methods typically fail to fully account for various properties of the grippers, such as their shape, actuation capability, and opening angle \cite{lee2015hierarchical, cruciani2018dexterous}. Consequently, they often generate motions that are difficult to realize in physical execution, such as rotating an object while one of its fulcrums slides on a planar surface \cite{lee2015hierarchical}. 

To address these limitations and ensure the physical realizability of manipulation actions, we develop a gripper-aware, library-driven hierarchical planner (LDHP). Concretely, we introduce a primitive library tailored to a parallel-jaw gripper and couple it with a two-tier search: a top-tier contact-state graph plans an object-pose sequence using MoveObject primitives, and a bottom-tier grasp graph verifies executable grasp sequences across poses using AdjustGrasp primitives. When contact-rich segments are fragile (e.g., tip), we bisect the segment to recover feasibility, yielding end-to-end plans that are physically consistent and immediately executable on hardware. This design addresses the practical pain points above. First, execution uses only low-dimensional force/torque feedback to gate contact events-avoiding reliance on multi-camera rigs or heavy tracking, while retaining closed-loop safeguards during contact transitions. Second, by checking collision and quasi-static stability under Coulomb friction, the planner naturally adapts to changes in physical conditions without retraining.

We validate the framework on a robot arm with different parallel-jaw gripper settings, across various tasks, including lifting an object from the table without grasping, and inserting the object into a slot via biplanar sliding and pushing. We also further validate generalization to different shapes and modified environments, and conduct ablations that confirm the necessity of our proposed primitives. The main contributions of this article are as follows:

\begin{enumerate}
    \item A gripper-aware primitive library that captures what a parallel-jaw gripper can physically execute, enabling implementable non-prehensile actions.
    \item A hierarchical planning algorithm that fuses a contact-state graph for object-pose planning with a grasp graph for executable regrasp sequences, plus a bisection refinement to recover feasibility on contact-sensitive segments.
    \item Real-world validation with various long-horizon dexterous tasks, including generalization tests and ablation.
\end{enumerate}

\section{RELATED WORK}

\textbf{Non-prehensile dexterous manipulation:}
Non-prehensile dexterous manipulation reorients or repositions objects without relying on rigid force-closure grasps. Instead, it leverages environmental constraints (e.g., table edges, slot walls), contact mechanics (friction, rolling/sliding), and external forces to maintain stability. In contrast to traditional dexterous in-hand manipulation that typically relies on multi-finger, multi-DOF hands and demonstration-heavy pipelines~\cite{bai2014dexterous, mnyusiwalla2015new, yuan2020design}, a growing body of work achieves in-hand or near-hand dexterity with simple low-DOF parallel-jaw grippers by exploiting the environment. Representative examples include extrinsic dexterity and environment-aided strategies for picking, repositioning and shallow insertion\cite{dafle2014extrinsic, chavan2020planar, kim2019shallow, he2021scooping}. Our approach follows this line by explicitly encoding gripper-aware manipulation primitives that harness environmental contact under quasi-static assumptions.

\textbf{Manipulation planning:}
Sampling-based methods~\cite{a,b} accelerate search in high-dimensional spaces by sampling feasible states on constraint manifolds and pruning invalid trajectories; while they scale well in kinematics, they typically omit explicit contact-mode reasoning, which limits reliability in NPDM regimes where stability hinges on precise contact evolution (e.g., tipping or controlled sliding). A second family models gripper–object–environment interactions explicitly so that feasibility arises from contact mechanics. Graph search over discrete contact configurations has been used to plan extrinsic dexterity~\cite{dafle2014extrinsic} and motion-cone–based sliding~\cite{chavan2020planar}. Two-tiered formulations coordinate object trajectories with contact locations~\cite{l}, yet often abstract away real gripper geometry and opening constraints, impeding physical validation. Extensions from 2D mode planning to 3D have been demonstrated~\cite{i}, but frequently assume sticking-only contacts, oversimplifying the richer mechanics that include controlled rolling, sliding, and contact breaking. Other approaches that enumerate contact configurations while ignoring object geometry~\cite{j} are prone to collision and execution failures. Hybrid schemes combine a discrete logic over interaction modes with continuous trajectory optimization to capture richer interactions such as hitting and throwing~\cite{c}. Learning-based paradigms—including reinforcement learning (RL)~\cite{g} and learning from demonstration~\cite{p}—pursue adaptability via policies trained in simulation or from demonstrations; however, their success depends strongly on simulation fidelity, and unmodeled effects (friction variation, wear, geometric tolerances) often degrade real-world repeatability, especially for precision insertion.


\begin{figure}[!t]
    \centering
    \includegraphics[width=0.9\linewidth]{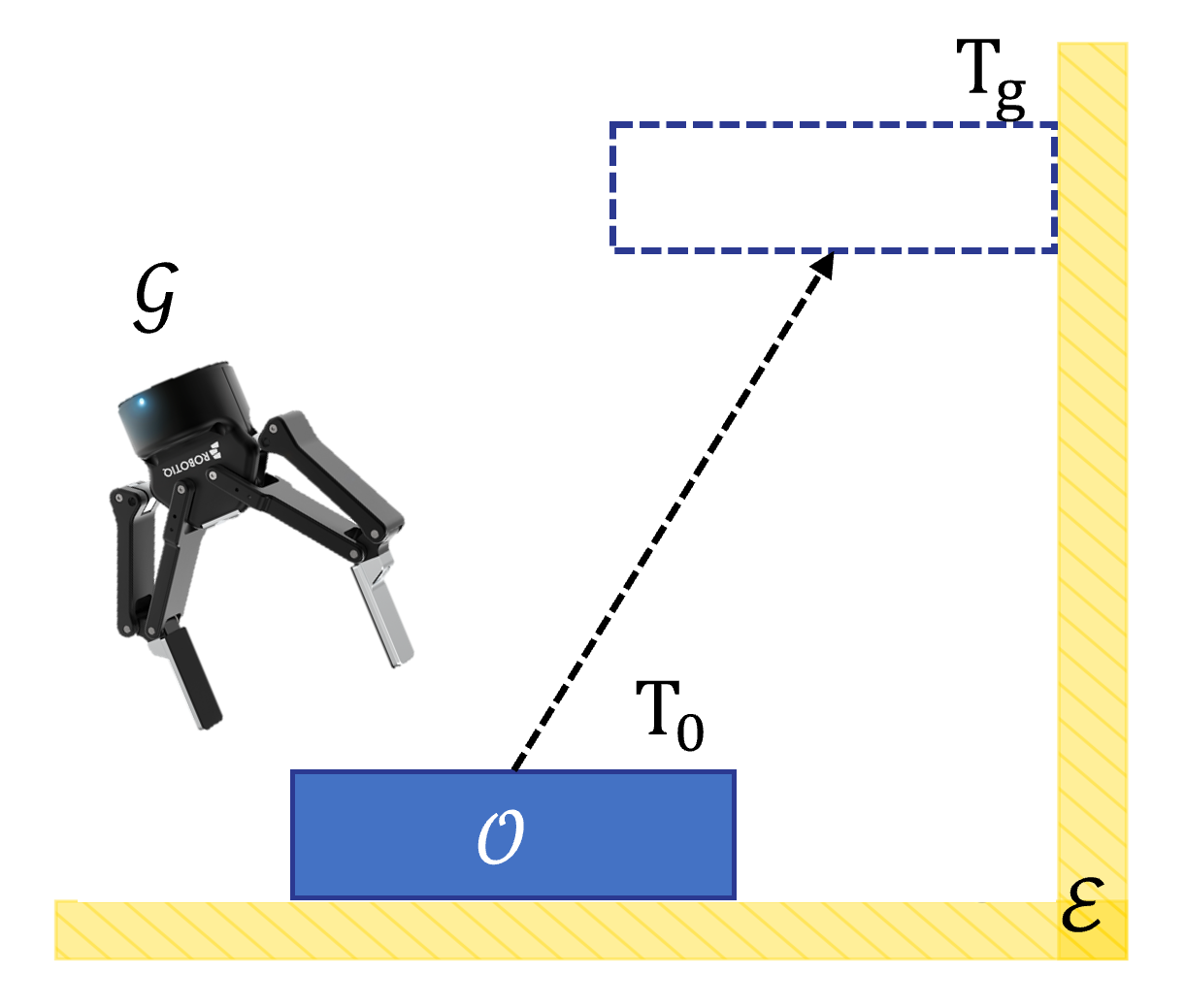}
    \vspace{-10pt}
    \caption{\textbf{Problem formulation.} A rigid object $\mathcal{O}$ with initial pose $T_0$ is placed inside a static environment $\mathcal{E}$ and manipulated by a parallel-jaw gripper $\mathcal{G}$. The task is to compute a feasible sequence of non-prehensile motion primitives that moves the object to the goal pose $T_g$.}
    \label{fg2}
\vspace{-18pt}
\end{figure}

\section{PROBLEM DESCRIPTION}

We formulate the problem of non-prehensile dexterous manipulation as finding a feasible sequence of object poses and associated grasps that transforms an object from an initial state to a desired goal, subject to the mechanics of contact and the physical constraints of the gripper.

Let the object be denoted by $\mathcal{O}$, with configuration space $\mathcal{T} \subset SE(2)$ since we restrict to planar quasi-static motions. An object state is written as $T \in \mathcal{T}$, with the initial and goal states given by $T_0$ and $T_g$ (Fig. \ref{fg2}). The environment $\mathcal{E}$ is modeled as a static rigid structure composed of polyline boundaries, which interact with the object during manipulation. The gripper is denoted by $\mathcal{G}$, where a grasp $g \in \mathcal{G}$ is defined by the gripper pose, opening width, and contact mode with the object. A manipulation plan is a sequence  
\begin{equation}
\pi = \big[ (T_0,g_0) \;\xrightarrow{m_1}\; (T_1,g_1) \;\; \cdots \;\xrightarrow{m_K}\; (T_K,g_K) \big],
\end{equation}  
where each edge corresponds to executing a motion primitive $m_k \in \mathcal{M}$ drawn from a finite primitive library. Each primitive specifies a local transformation of the object configuration: 
\begin{equation}
T_{k+1} = F_{m_k}(T_k),
\end{equation}  
subject to its parameterization (e.g., slide direction, pushing location, or pivot angle).  

For each transition $(T_k,g_k) \xrightarrow{m_k} (T_{k+1},g_{k+1})$ to be valid, several conditions must hold. First, there must exist feasible contact forces $\{f_i\}$ satisfying Coulomb friction cones and static moment equilibrium, so that the motion is quasi-statically executable. Second, the grasp $g_k$ must respect the gripper’s geometric constraints, including opening width, approach direction, and collision avoidance with the object $\mathcal{O}$ and environment $\mathcal{E}$. Third, the transition from $g_k$ to $g_{k+1}$ must admit a collision-free path in the arm’s configuration space. Finally, the intermediate object pose $T_{k+1}$ must lie in a stable region of $\mathcal{T}$, such that the object remains supported and does not topple or slip.  

These constraints define a contact-state graph $\mathcal{C} = (\mathcal{V}, \mathcal{E})$, corresponds to a contact state (object–environment relation) \cite{xiao2001automatic} without grasp, and each edge $(v_i,v_j) \in \mathcal{E}$ represents a valid primitive execution. Planning thus involves finding a path in $\mathcal{C}$ from an initial state $(T_0, g_0)$ to a goal state $(T_g, g)$ for a feasible grasp $g$.

In summary, the task is to synthesize a path on the contact-state graph that satisfies quasi-static feasibility, grasp realizability, collision-free reachability, and stability at all intermediate states. The central challenge is that the size of $\mathcal{C}$ grows combinatorially with the number of possible grasps and primitive instantiations, requiring structured decomposition and efficient search strategies to produce executable plans.  

\section{METHODOLOGY}
\label{sec:method}

\begin{figure}[!t]
    \centering
    \includegraphics[width=1\linewidth]{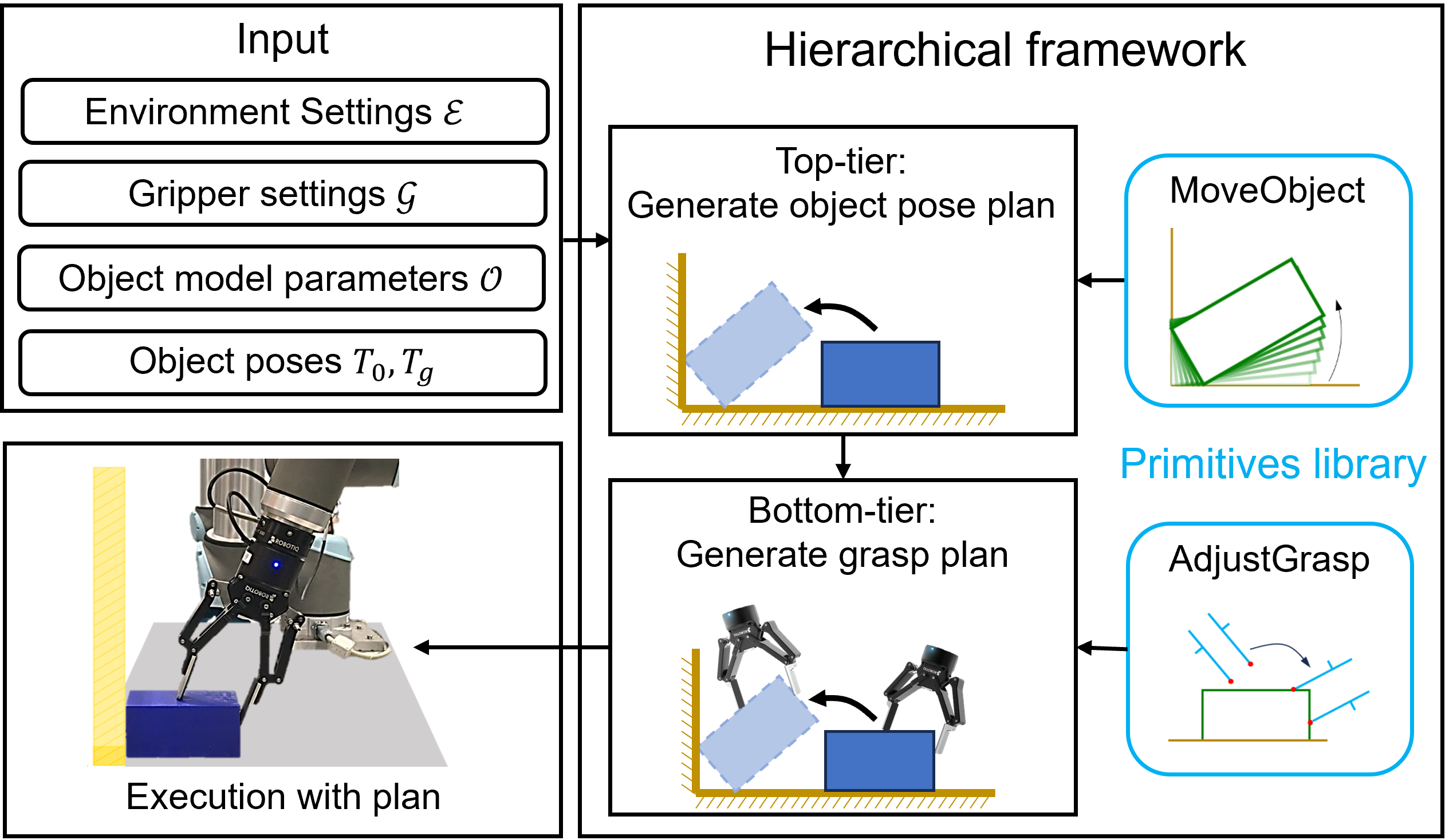}
    \caption{\textbf{System overview.} From left to right: inputs (environment $\mathcal{E}$, gripper $\mathcal{G}$, object model $\mathcal{O}$, and initial/goal poses $T_0,T_g$); the hierarchical framework that (top tier) generates an object-pose plan using \emph{MoveObject} primitives and (bottom tier) generates a grasp plan using \emph{AdjustGrasp} primitives; and execution of the resulting plan on the robot.}
    \label{fg3}
\vspace{-20pt}
\end{figure}

This section details the proposed Library-Driven Hierarchical Planner (LDHP). 
The pipeline couples a \emph{top-tier} contact-state planner that proposes an object-pose trajectory using MoveObject primitives with a \emph{bottom-tier} grasp planner that verifies and synthesizes an executable grasp sequence using AdjustGrasp primitives; when contact-sensitive segments (e.g., tip, biplanar-slide) fail grasp feasibility, a dichotomy refinement recovers feasibility. See the system overview in Fig.~\ref{fg3}. The full procedure is summarized in Algorithm~\ref{alg:flow-chart}.

\subsection{Inputs}
\label{subsec:inputs} 
We consider planar quasi-static manipulation of a rigid object $\mathcal{O}$ inside a static environment $\mathcal{E}$ using a parallel-jaw gripper $\mathcal{G}$. 
The object configuration space is $\mathcal{T}\subset SE(2)$; the task is to move $\mathcal{O}$ from an initial pose $T_0\in\mathcal{T}$ to a target pose $T_g\in\mathcal{T}$. 
The planner therefore receives the tuple
\[
\mathbf{I} = \big(\mathcal{O},\,\mathcal{E},\,\mathcal{G},\,T_0,\,T_g,\,\{\mu_i\}\big),
\]
which fully specifies the geometric and physical constraints used throughout the pipeline.

\paragraph{Environment \(\mathcal{E}\)}
The environment is modeled as a set of rigid polyline boundaries that constrain feasible motions. 
We represent its geometry by vertices $\mathcal{V}_E=\{v^E_1,\dots,v^E_{N_E}\}$ and straight-line edges $\mathcal{E}_E=\{e^E_1,\dots,e^E_{M_E}\}$ connecting adjacent vertices, yielding a piecewise-linear boundary on which object–environment contacts are formed. 
This representation directly supports contact-state reasoning (which edge(s) are in contact) and collision checks along planned motions.

\begin{figure}[!t]
    \includegraphics[width=1\linewidth]{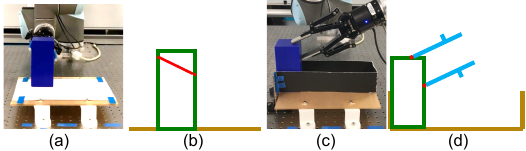}
    \caption{\textbf{Gripper configurations used in this work.}
    (a--b) \textbf{\emph{Configuration~I}:} two-finger gripper with cylindrical flanks, modeled as a two-point contact abstraction with fixed inter-finger distance.
    (c--d) \textbf{\emph{Configuration~II}:} cuboid fingertips represented by polyline contours.
    A grasp is valid only if a collision-free gripper pose exists for approach/retreat and for any required regrasp under opening-width limits.}
    \label{fg4}
\vspace{-20pt}
\end{figure}

\paragraph{Gripper \(\mathcal{G}\)}
The gripper model captures finger geometry and opening limits. 
We use two practically relevant configurations also shown in Fig.~\ref{fg4}: 
Configuration~I abstracts cylindrical flank-contact fingers as a two-point model with fixed inter-finger distance; 
Configuration~II abstracts cuboid fingertips as a polyline contour that must remain collision-free with $\mathcal{O}$ and $\mathcal{E}$ during approach and regrasp. 
Both configurations share the same kinematic frame definition and will be used interchangeably depending on hardware.

\paragraph{Object \(\mathcal{O}\)}
The object shape is given by vertices $\mathcal{V}_O=\{v^O_1,\dots,v^O_{N_O}\}$ and straight-line edges  $\mathcal{L}_O=\{l^O_1,\dots,l^O_{N_O}\}$ defining its boundary; 
the center of mass is $o\in\mathbb{R}^2$ in the object frame and is used for quasi-static moment balance checks. 
All contact computations (collision, mode identification, and stability) use this polyline model.

\paragraph{Poses \(T_0,T_g\)}
The initial and goal poses belong to $\mathcal{T}\subset SE(2)$ and are specified in the world frame. 
When either pose is contact-free, our planner will prepend/append a short free-space segment (\textsc{air-move}) to connect the pose to a nearby contact state, enabling contact-state planning downstream.

\paragraph{Friction \(\{\mu_i\}\)}
Each potential contact interface (object edge, environment edge, fingertip) carries an indexed Coulomb friction coefficient $\mu_i$. 
These coefficients parameterize the friction cones used in contact-mode feasibility tests and moment labeling along candidate motions; heterogeneous values are allowed to reflect real surfaces.

With $(\mathcal{O},\mathcal{E},\mathcal{G})$ fixing geometry and hardware, $(T_0,T_g)$ defining the task, and $\{\mu_i\}$ setting contact mechanics, the input tuple $\mathbf{I}$ determines the search spaces for both tiers of our planner as detailed in Sec.~\ref{sec:method}.

\subsection{Primitive Library}
\label{subsec:library}
\begin{figure}[!t]
    \includegraphics[width=1\linewidth]{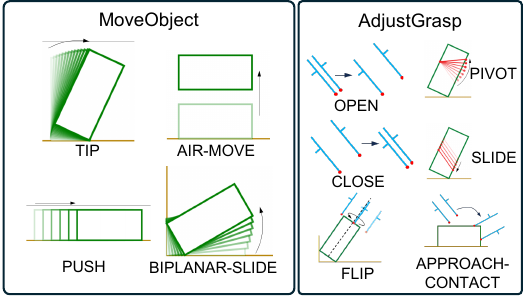}
    \caption{\textbf{Visualization of MoveObject and AdjustGrasp primitives.}}
    \label{fg5}
\vspace{-18pt}
\end{figure}

We instantiate a gripper-aware primitive library
$\mathcal{M}=\mathcal{M}^{\mathrm{MO}}\cup\mathcal{M}^{\mathrm{AG}}$:
\emph{MoveObject} ($\mathcal{M}^{\mathrm{MO}}$) changes the object pose while keeping the current grasp, and
\emph{AdjustGrasp} ($\mathcal{M}^{\mathrm{AG}}$) modifies the grasp at a fixed pose, as shown in Fig. \ref{fg5}.
Each primitive $m$ is a typed operator with parameters $\theta$ and a state transformer $F_m$,
\[
\begin{cases}
T' = F_m(T,\theta), & m\in\mathcal{M}^{\mathrm{MO}},\\[2pt]
g' = F_m(g,\theta;T), & m\in\mathcal{M}^{\mathrm{AG}}.
\end{cases}
\]
Feasibility is evaluated using the procedure described at the end of this subsection (see ``Feasibility check'').

\paragraph{MoveObject ($\mathcal{M}^{\mathrm{MO}}$)}
We use four gripper-aware primitives to traverse contact states: \textsc{tip}, \textsc{push}, \textsc{biplanar-slide}, and \textsc{air-move}. Their semantics and parameters are detailed in Table~\ref{tab:moveobject} and Fig.~\ref{fg5}.

\paragraph{AdjustGrasp ($\mathcal{M}^{\mathrm{AG}}$)}
We use \textsc{open}, \textsc{close}, \textsc{pivot}, \textsc{slide}, \textsc{flip}, and \textsc{approach-contact} to establish, switch, and refine grasps around $\mathcal{O}$; see Table~\ref{tab:adjustgrasp} and Fig.~\ref{fg5} for details.

\paragraph{Feasibility check for a primitive instance.}
For either family, we discretize the motion into intermediate states and perform:
(1) collision checks between the gripper model (Config.~I/II in Fig.~\ref{fg4}), the object $\mathcal{O}$ and the environment $\mathcal{E}$;
(2) contact-mode identification (sticking/rolling/sliding/separation) at all active interfaces; and
(3) a quasi-static check that there exist contact reactions $\{f_i\}$ within Coulomb cones such that net force and moment about the center of mass vanish.
If all samples pass, the instance is accepted. For contact-sensitive \textsc{tip}/\textsc{biplanar-slide} segments, dichotomy refinement is used when bottom-tier grasp planning fails (Sec.~\ref{subsec:two-tier}).

\subsection{Candidate Grasp Sets}
\label{subsec:cand-grasps}
Given a gripper model $\mathcal{G}$ and an object $\mathcal{O}$, we construct grasp candidates in three steps: (i) enumerate grasps in the object frame; (ii) filter them by the current object pose $T$ to obtain pose-conditioned grasps; and (iii) validate, for each MoveObject segment $m\in\mathcal{M}^{\mathrm{MO}}$, which grasps can execute $T\!\to\!T'$ under quasi-static mechanics. We also model intra-pose transitions between different grasps using AdjustGrasp primitives.

\paragraph{Enumeration in the object frame}
We discretize each object edge into $K$ contact samples (we use $K{=}100$ in all experiments). 
A grasp is represented by a pair of fingertip contacts $g=\{f_x,f_y\}$ on the polygon boundary of $\mathcal{O}$ in the object frame. 
For Configuration~I (two-point abstraction), admissible $g$ must satisfy the fixed inter-finger distance. 
For Configuration~II (polyline fingertips), a grasp is kept only if there exists at least one collision-free gripper pose realizing the contact pair in the object frame. 
We augment the set with a special symbol $\{0,0\}$ to denote the no-contact state used during \textsc{open} and \textsc{approach-contact}.
Let $\mathcal{G}$ denote the resulting discrete grasp set. 

\paragraph{Pose-conditioned grasps}
At a world pose $T\in\mathcal{T}$, we transform each $g\in\mathcal{G}$ into the world frame and discard those that collide with $\mathcal{E}$ or violate opening-width limits; the survivors form
\[
\mathcal{G}_T
= \bigl\{\, g\in\mathcal{G}\,\bigm|\,
\substack{\text{collision-free gripper pose at }T\\
\text{and within opening-width limits}}
\bigr\}.
\]

\paragraph{Motion-conditioned grasps for a segment}
Consider a MoveObject instance $m:T\!\to\!T'$ with parameters $\theta$. 
A grasp $g$ is admissible for $m$ if (i) $g\in \mathcal{G}_T\cap \mathcal{G}_{T'}$ and (ii) there exists a quasi-statically feasible contact evolution along the interpolated motion. 
We discretize $m$ into intermediate poses $\{T^{(j)}\}_{j=0}^J$ and, at each $T^{(j)}$, (a) check collisions for the gripper realizing $g$, (b) identify the contact mode (sticking/rolling/sliding/separation) at all active interfaces, and (c) test whether there exist contact reactions $\{f_i^{(j)}\}$ inside Coulomb friction cones such that net force and moment about the center of mass vanish. 
If all samples pass, $g$ is accepted for this segment. 
The admissible set is
\[
\mathcal{G}_m
= \bigl\{\, g \in \mathcal{G}_T \cap \mathcal{G}_{T'} \,\bigm|\,
\substack{\text{collision-free along } m\\
\text{and quasi-static feasible}}
\bigr\}.
\]

\begin{table}[!t]
\centering
\caption{MoveObject primitives and parameterization}
\label{tab:moveobject}
\small
\begin{tabular}{p{0.20\linewidth} p{0.73\linewidth}}
\toprule
\textbf{Primitive} & \textbf{Semantics, Parameters, Preconditions} \\
\midrule
\textsc{tip} & Rotate $\mathcal{O}$ about a sticking/rolling contact on an environment edge. 
Params: pivot location on edge, rotation sense, angle $\varphi$. 
Pre: a feasible contact exists at $T$; post: same grasp $g$, updated $T'$. \\
\textsc{push} & Translate along a single environment edge with maintained contact.
Params: tangential direction and travel $d$ along the edge. 
Pre: contact at $T$ and admissible friction; post: same $g$, updated $T'$. \\
\textsc{biplanar-slide} & Simultaneous sliding on two environment edges (e.g., corner guidance), enabling coupled translation--rotation.
Params: orientation change $\Delta\theta$ and travel.
Pre: two admissible contacts at $T$; post: same $g$, updated $T'$. \\
\textsc{air-move} & Free-space motion without environment contact.
Params: waypoints in $\mathcal{T}$.
Pre: collision-free path exists with same $g$; post: updated $T'$. \\
\bottomrule
\end{tabular}
\label{tab1}
\vspace{-18pt}
\end{table}

\paragraph{Intra-pose transitions between grasps}
At a fixed pose $T$, two grasps $g_i,g_j\in\mathcal{G}_T$ are connected if there exists a collision-free sequence of AdjustGrasp primitives that transforms $g_i$ to $g_j$ while respecting contact-mode feasibility; typical sequences include \textsc{pivot}, \textsc{slide}, \textsc{flip}, or routes through $\{0,0\}$ using \textsc{open}/\textsc{approach-contact}. 
These transitions define edges inside the grasp graph used by the bottom tier.

\paragraph{Grasp quality and edge costs}
For $g$ admissible on a motion $m$, we assign a quality
\[
S(g,m) \;=\; w_1\, s(g,m) \;+\; w_2\, \alpha(g,m) ,
\]
where $s\in\{0,1\}$ indicates force-closure violation ($0$ means force-closure holds and vice versa), and $\alpha$ is the average angle between the contact reaction and the contact normal over the sticking/rolling contacts along $m$ (smaller is better). 
Inter-pose edges $(g,T)\!\to\!(g,T')$ receive cost $S_g$; intra-pose AdjustGrasp edges are weighted by a combination of finger travel and primitive switches. 
These costs bias the search toward stable, short, and executable grasp plans.

\begin{table}[!t]
\centering
\caption{AdjustGrasp primitives and preconditions}
\label{tab:adjustgrasp}
\small
\begin{tabular}{p{0.20\linewidth} p{0.73\linewidth}}
\toprule
\textbf{Primitive} & \textbf{Semantics, Parameters, Preconditions} \\
\midrule
\textsc{open}/\textsc{close} & Change opening width to disengage/engage contacts.
Params: opening $\Delta w$.
Pre: collision-free finger motion; respects opening-width limits. \\
\textsc{pivot} & Rotate the gripper about a contacting finger to change orientation.
Params: pivot finger and angle.
Pre: at least one sticking/rolling finger contact. \\
\textsc{slide} & Slide a contacting finger along an object edge to relocate contact.
Params: edge id and arc-length.
Pre: sliding along the edge, no collision. \\
\textsc{flip} & 180$^\circ$ in-plane rotation of the gripper about its axis to switch facing.
Params: rotation sense.
Pre: collision-free clearance during rotation. \\
\textsc{approach-contact} & Establish a two-finger contact from free space.
Params: approach pose and path.
Pre: collision-free approach and reachable opening. \\
\bottomrule
\end{tabular}
\label{tab2}
\vspace{-18pt}
\end{table}

\begin{algorithm}[!t]
  \caption{Hierarchical Manipulation Planning Framework}
  \label{alg:flow-chart}
  \begin{algorithmic}[1]
    \REQUIRE Object $\mathcal{O}$, environment $\mathcal{E}$, gripper model $\mathcal{G}$ (geometry, opening limits), friction coefficients $\{\mu_i\}$, initial and goal poses $T_0, T_g \in \mathrm{SE}(2)$, primitive library $\mathcal{M}$
    \ENSURE Feasible plan $\pi = \big[(T_0,g_0)\!\xrightarrow{m_0}\!(T_1,g_1)\!\xrightarrow{m_1}\!\cdots\!\xrightarrow{m_{n-1}}\!(T_n{=}T_g,g_n)\big]$

    \STATE Initialize top-tier iteration $h_{\rm iter} \gets 0$
    \WHILE{True}
      \STATE $h_{\rm iter} \gets h_{\rm iter} + 1$
      \IF{$h_{\rm iter} > H_{\max}$}
        \STATE \textbf{throw exception}: \textit{Feasible plan unavailable}
        \STATE \textbf{break}
      \ENDIF

      \STATE // Top-tier: Generate object pose trajectory
      \STATE Let $\mathbf{T}_{0:n} = \big[T_0 \xrightarrow{m_0} T_1 \dots \xrightarrow{m_{n-1}} T_n{=}T_g\big]$ be the pose trajectory computed by the top-tier, with associated primitive list $\{m_k\}_{k=0}^{n-1}$, where each $m_k$ denotes the motion segment between $T_k$ and $T_{k+1}$.
      \IF{top-tier planning fails (no valid $\mathbf{T}_{0:n}$)}
        \STATE \textbf{continue} \text{ to next top-tier planning iteration}
      \ENDIF

      \STATE // Bottom-tier: Solve grasp feasibility for $\mathbf{T}_{0:n}$
      \STATE Initialize bottom-tier iteration counter $l_{\rm iter} \gets 0$
      \WHILE{true}
        \IF{$l_{\rm iter} > L_{\max}$}
          \STATE \textbf{break} \text{ (back to top-tier re-planning)}
        \ENDIF

        \STATE // Compute grasp plan for current trajectory $\mathbf{T}_{0:n}$
        \STATE Let $\mathbf{g}_{0:n}$ be the grasp plan generated by the bottom-tier algorithm.
        \IF{$\mathbf{g}_{0:n}$ \text{ is feasible}}
          \STATE \textbf{return} $(\mathbf{T}_{0:n}, \mathbf{g}_{0:n})$ \text{ as the solution}
        \ELSIF{$m_k$ \text{ is \textbf{tip}/\textbf{biplanar-slide} motion}}
          \STATE \textbf{// Refine motion segments via dichotomy}
          \STATE Split $m_{k^\ast}$ into two sub-segments $m_{k^\ast}^{(1)}$, $m_{k^\ast}^{(2)}$ and update $\mathbf{T}_{0:n}$ and $\{m_k\}$ accordingly.
          \STATE $l_{\rm iter} \gets l_{\rm iter} + 1$
          \STATE \textbf{continue} \text{ to next bottom-tier iteration}
        \ELSE
          \STATE \textbf{break} \text{ (back to top-tier re-planning)}
        \ENDIF
      \ENDWHILE
    \ENDWHILE
  \end{algorithmic}
\label{alg:flow-chart}
\end{algorithm}

\subsection{Two-Tier Planning}
\label{subsec:two-tier}

Given the inputs in Sec.~\ref{subsec:inputs} and the primitive library in Sec.~\ref{subsec:library}, our planner searches over two coupled structures: a top-tier contact-state graph that proposes an object-pose path using MoveObject primitives, and a bottom-tier grasp graph that verifies and synthesizes an executable grasp sequence using AdjustGrasp primitives. The tiers communicate through motion-conditioned grasp sets $\mathcal{G}_{m_k}$ defined in Sec.~\ref{subsec:cand-grasps}; when these sets are empty for contact-sensitive segments, a dichotomy refinement is triggered. Algorithm~\ref{alg:flow-chart} provides the pseudocode of this two-tier procedure.

\paragraph{Top tier, object-pose planning on a contact-state graph}
\textbf{Step 1:} Identify nearby \emph{contact poses} $T'_0$ and $T'_g$ associated with $T_0$ and $T_g$, so that subsequent reasoning occurs in contact states. In our work, $T'_0$ and $T'_g$  are selected to be translated along horizontal or vertical directions from $T_0$ and $T_g$ without environmental collisions.
\textbf{Step 2:} Build a contact-state graph in accordance with the algorithms detailed in \cite{xiao2001automatic} whose nodes encode object--environment contact relations (which edge(s) of $\mathcal{E}$ are active, stick/roll status) and whose directed edges are parameterized \emph{MoveObject} instances $m\!\in\!\mathcal{M}^{\mathrm{MO}}$ that are kinematically admissible between the incident states. Running Dijkstra from the state of $T'_0$ to that of $T'_g$ yields a discrete state sequence $\sigma=\{s_0,\ldots,s_M\}$. 
\textbf{Step 3:} \emph{Assemble the object-pose path via three sub-steps:}
\emph{(Sub-step~1)} for each transition $s_k\!\to s_{k+1}$, sample primitive parameters and integrate them to produce a local pose subpath $\widehat{\mathbf{T}}_{k:k+1}$ with endpoints $T_k$ and $T_{k+1}$, effectively following the chosen contact-state sequence;
\emph{(Sub-step~2)} within the terminal contact state $s_M$, execute a short in-state motion to align the object with the target contact pose $T'_g$; 
\emph{(Sub-step~3)} append a free-space \textsc{air-move} from $T'_g$ to the exact goal $T_g$ (and symmetrically from $T_0$ to $T'_0$ if needed). Concatenating all subpaths gives $\mathbf{T}_{0:n}=[T_0,\ldots,T_n{=}T_g]$ with associated primitives $\{m_k\}_{k=0}^{n-1}$, where $T_{k+1}=F_{m_k}(T_k)$.

\paragraph{Bottom tier, grasp-sequence planning on a layered grasp graph}
Given $\mathbf{T}_{0:n}$ and $\{m_k\}$, we compute pose-conditioned grasp sets $\mathcal{G}_{T_k}$ for all $k$ and motion-conditioned sets $\mathcal{G}_{m_k}$ for all segments (Sec.~\ref{subsec:cand-grasps}). 
We then construct a layered graph whose $k$-th layer contains nodes $(T_k,g)$ for all $g\!\in\!\mathcal{G}_{T_k}$. 
Edges of two types are added: (i) \emph{inter-pose} edges $(T_k,g)\!\to\!(T_{k+1},g)$ for $g\!\in\!\mathcal{G}_{m_k}$, with cost
\[
c^{\mathrm{inter}}(k,g)=S_g(m_k),
\]
where $S_g$ is the grasp quality defined in Sec.~\ref{subsec:cand-grasps}; and (ii) \emph{intra-pose} edges $(T_k,g_i)\!\to\!(T_k,g_j)$ when there exists a collision-free sequence of AdjustGrasp primitives that realizes $g_i\!\to\!g_j$ at $T_k$, weighted by a combination of finger travel and primitive switches,
\[
c^{\mathrm{intra}}(k;i\!\to\!j)=\lambda_1\,\text{travel}(g_i,g_j;T_k)+\lambda_2\,\text{switches}(g_i,g_j).
\]
A shortest path on this grasp graph (Dijkstra) returns $\mathbf{g}_{0:n}$; together with $\mathbf{T}_{0:n}$ this yields an executable plan $\pi$.

\paragraph{Dichotomy refinement and backtracking}
If some segment $m_k$ admits no motion-conditioned grasp, i.e., $\mathcal{G}_{m_k}=\emptyset$, and $m_k\!\in\!\{\textsc{tip},\textsc{biplanar-slide}\}$, we split $m_k$ into two sub-motions by bisecting its parameter range (e.g., halving the rotation or travel), insert the new poses into $\mathbf{T}_{0:n}$, and recompute the affected sets $\mathcal{G}_{T}$ and $\mathcal{G}_{m}$ locally. 
This refinement repeats up to $L_{\max}$ times. 
If refinement still fails or $m_k$ is not in the sensitive set, we backtrack to the top tier to generate a new pose path (outer-loop limit $H_{\max}$).

\paragraph{Objective and completeness under discretization}
The overall objective is to minimize the summed edge costs on the grasp graph subject to feasibility on both tiers:
\[
\min_{\mathbf{T}_{0:n},\,\mathbf{g}_{0:n}} 
~\sum_{k=0}^{n-1} c^{\mathrm{inter}}(k,g_k)
~+~\sum_{k=0}^{n} \sum_{(g_i\to g_j)\in \mathcal{E}^{\mathrm{AG}}_k} c^{\mathrm{intra}}(k;i\!\to\!j),
\]
where $\mathcal{E}^{\mathrm{AG}}_k$ are the intra-pose AdjustGrasp edges used at $T_k$. 
Given finite parameter sampling on the top tier and finite discretization along motions for feasibility checks, the algorithm terminates with either a plan or \textsc{Failure} after at most $H_{\max}$ outer proposals and $L_{\max}$ refinements per sensitive segment.

We limit the number of parameter samples per state transition on the top tier and rank them by a lightweight heuristic (e.g., distance-to-goal and contact-change penalty); on the bottom tier we cache $\mathcal{G}_{T}$ and $\mathcal{G}_{m}$ to avoid repeated feasibility checks. 
Optional learned priors can be plugged in by reweighting $c^{\mathrm{inter}}$ with a predicted feasibility score $\hat p(g\,|\,m_k)$ or by prioritizing parameter samples on the top tier, without changing the planner’s interface.

\begin{figure*}[!t]
    \centering
    \includegraphics[width=\linewidth]{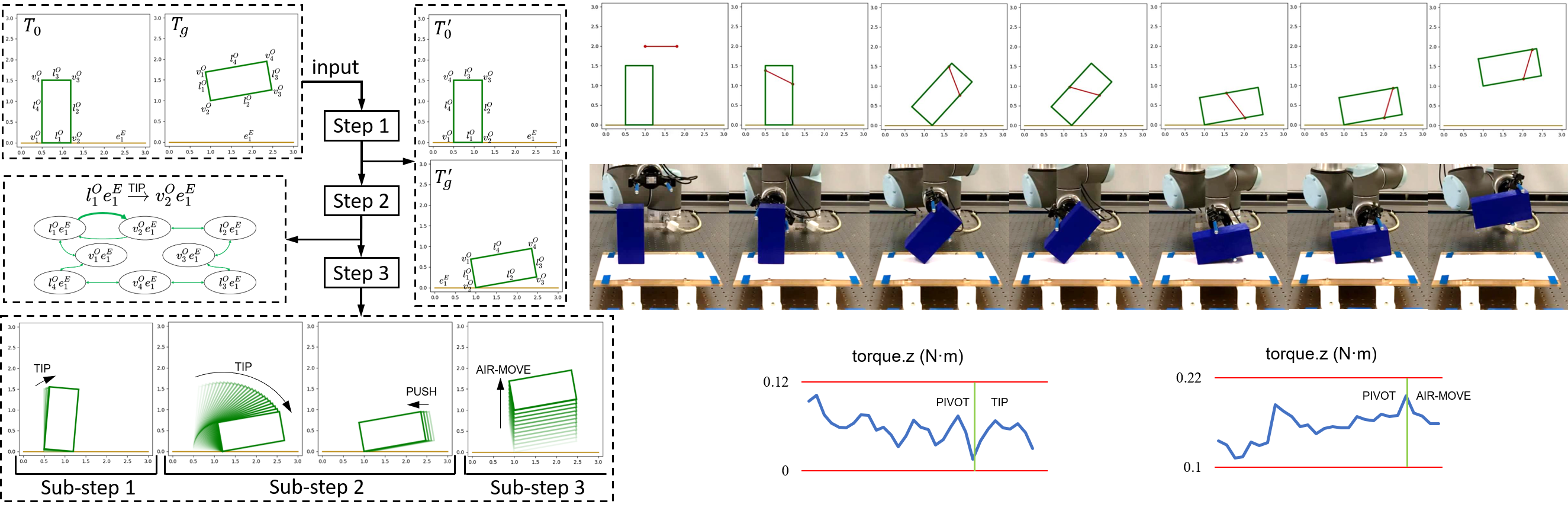}
    \put(-400,-3){\footnotesize (a)}
        \put(-155,114){\footnotesize (b)}
    \put(-155,56){\footnotesize (c)}
    \put(-210,-1){\footnotesize (d)}
    \put(-84,-1){\footnotesize (e)}
    \caption{\textbf{Task 1: zero-mobility lifting. }
    (a) Top-tier contact-state planning outputs; the selected plan is highlighted. 
    (b) Planned manipulation snapshots. 
    (c) Hardware snapshots. 
    (d,e) Wrist torque.z around the instant when both fingers contact the object (green line marks the instant).}
   \label{fig:fig6}
\vspace{-5pt}
\end{figure*}

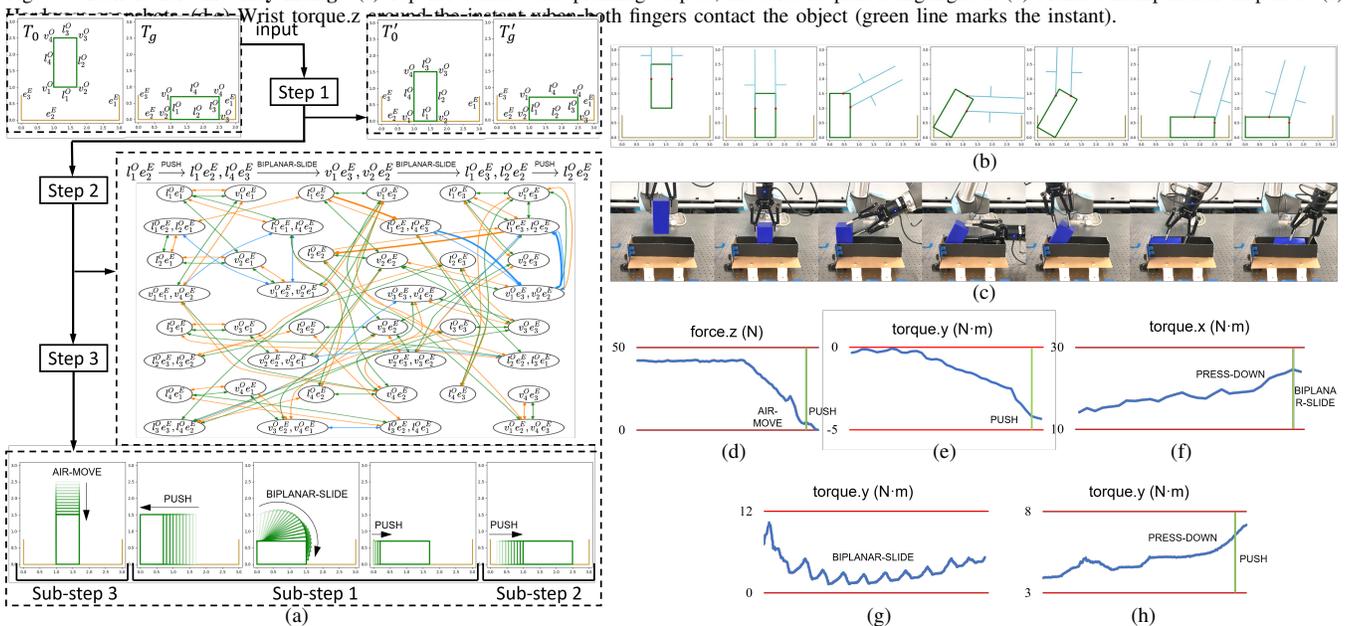
\begin{figure*}[!t]
    \centering
       \includegraphics[width=\linewidth]{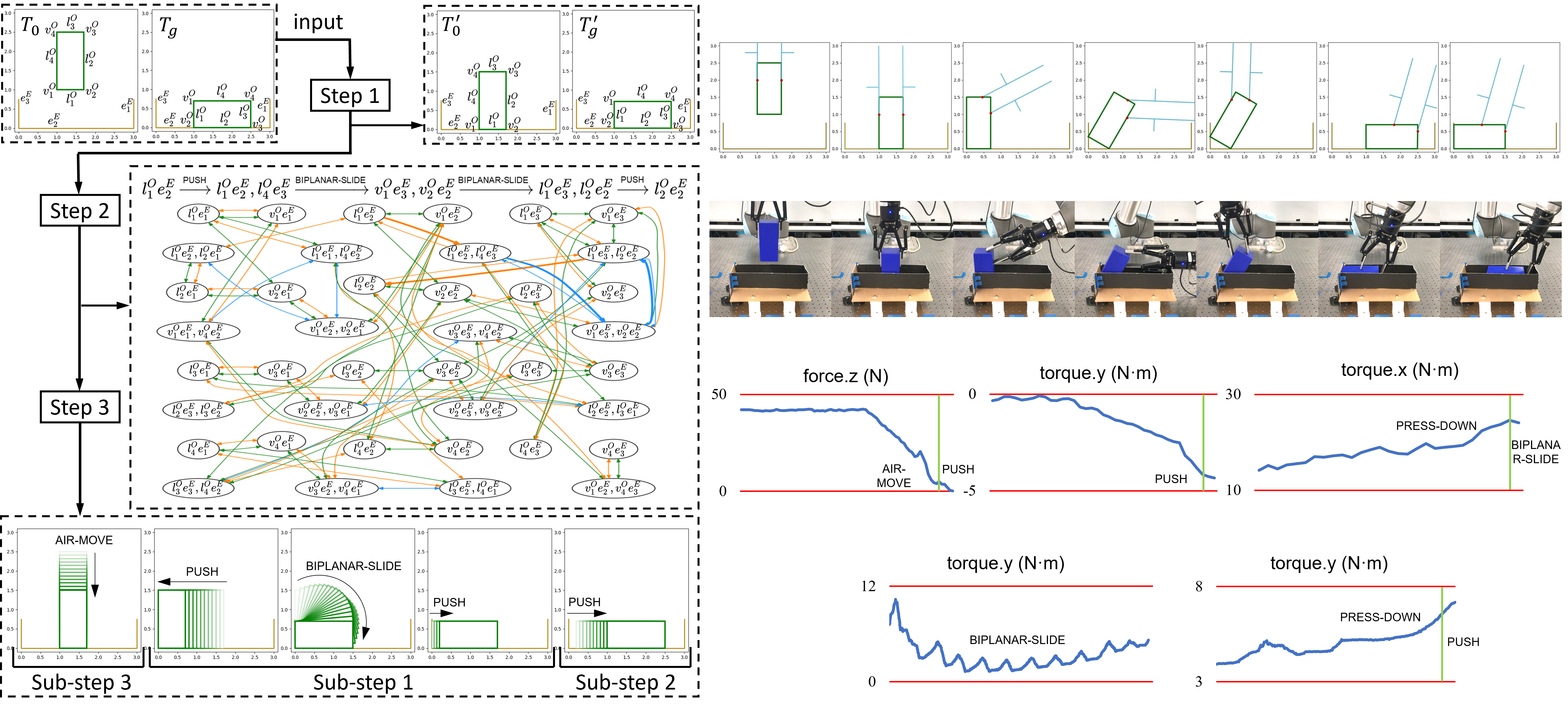}
       \put(-400,-2){\footnotesize (a)}
       \put(-140,170){\footnotesize (b)}
    \put(-140,120){\footnotesize (c)}
       \put(-235,60){\footnotesize (d)}
       \put(-155,60){\footnotesize (e)}
    \put(-65,60){\footnotesize (f)}
    \put(-80,-2){\footnotesize (h)}
    \put(-180,-2){\footnotesize (g)}
    \caption{\textbf{Task 2: slot insertion}. 
    (a) Top-tier contact-state search; green/orange/blue denote \textsc{tip}/\textsc{push}/\textsc{biplanar-slide}; the selected sequence is highlighted. 
    (b) Planned manipulation snapshots. 
    (c) Hardware snapshots. 
    (d–h) Force/torque traces used to gate key contact events.}
   \label{fig:fig7}
\vspace{-18 pt}
\end{figure*}

\section{EXPERIMENTS}
\label{sec:experiments}

We evaluate whether LDHP can turn contact-state plans into executable pose–grasp plans on real hardware under modest sensing.
Two representative case studies are considered: \textbf{Task 1} (zero-mobility lifting) and \textbf{Task 2} (slot insertion).
Across both, we reuse the same primitive library and planner settings—only object and environment geometry differ.
We then probe generalization to new shapes, altered environment dimensions, and inverse variants of the tasks.
Finally, an ablation study removes one primitive at a time to quantify which actions are indispensable for completing each task.

\subsection{Case I: Manipulation with Grippers of Zero Mobility}
\label{subsec:case1}
We first evaluate our framework under the zero-mobility gripper setting of \cite{mucchiani2018object}, where the opening cannot be adjusted (Configuration~I). The task is to pick up a $70{\times}150$\,mm rectangular object from a flat surface using a fixed $80$\,mm opening with a uniform friction coefficient $\mu=0.2$; Fig.~\ref{fig:fig6} summarizes the pipeline and snapshots.

Given $T_0$, $T_g$, and the ground edge $e^E_1$, the top tier first determines the contact poses $T'_0$ and $T'_g$. It then searches an $8$-node contact-state graph and returns the plan $l^O_1e^E_1\xrightarrow{\textsc{tip}}v^O_2e^E_1$ (Fig.~\ref{fig:fig6}a). The final pose path is assembled by concatenating three segments: a path that follows this contact-state sequence to reach $v^O_2e^E_1$, a short motion within that state to align with $T'_g$, and the free-space segment to $T_g$. On this pose path, the bottom tier builds a grasp graph with $1810$ nodes and extracts a shortest-path grasp sequence; the resulting manipulation plan is visualized in Fig.~\ref{fig:fig6}b and in the accompanying video. 

Experiments are conducted on a UR10 arm with a Robotiq~140 two-finger gripper (aluminum cylindrical fingers, radius $5$\,mm, initial spacing $80$\,mm) and a Robotiq FT-300 wrist force/torque sensor; snapshots are shown in Fig.~\ref{fig:fig6}c. Execution uses position control augmented with force/torque triggers: the \textsc{pivot} phase terminates when both fingers contact the object, detected by a threshold on torque.z (Fig.~\ref{fig:fig6}d–e). We repeat ten times and achieve success in all trials. 

Compared with the expert-designed finite-state machine of \cite{mucchiani2018object}, our planner autonomously decides when \textsc{pivot} is required based on the contact state and CoM geometry, instead of invoking it only after a failed lift attempt.

\subsection{Case II: Inserting an Object into a Slot}
\label{subsec:case2}
We further evaluate the framework on a slot-insertion task using Configuration~II. A $70{\times}150$\,mm rectangular object is initially held above a $300{\times}75$\,mm slot; the goal is to insert the object so that it finally rests horizontally in the slot under quasi-static constraints with $\mu_e=0.03$ and $\mu_g=0.6$, where $\mu_e$ ($\mu_g$) is the friction coefficient between the object and the environment (gripper) (leftmost panel of Fig.~\ref{fig:fig7}c). 

The top tier proceeds as before. Fig.~\ref{fig:fig7}a summarizes the outputs of each step; in Step~2 the contact-state search explores forty candidate states. Transitions feasible by \textsc{tip}, \textsc{push}, and \textsc{biplanar-slide} are encoded by green, orange, and blue arrows, respectively, and the selected sequence is highlighted by thick arrows. Sampling along this sequence yields the pose path passed to the bottom tier. Given this pose path, the bottom tier constructs a grasp graph with $13321$ nodes and computes a shortest-path grasp plan, shown in Fig.~\ref{fig:fig7}b and in the video. 

Validation is carried out on the same UR10 and Robotiq~140 platform as in Case~I (only the fingers differ), using the same object (Fig.~\ref{fig:fig7}c). Execution is gated by force/torque triggers: the robot descends until the object touches the slot bottom and then translates toward the slot wall until contact (Fig.~\ref{fig:fig7}d–e); it subsequently adjusts the grasp and performs a \textsc{biplanar-slide}, with force/torque feedback ensuring adequate environmental contact to reduce contact loss (i.e., press-down, Fig.~\ref{fig:fig7}f–g). When the object has rotated by a preset angle, \textsc{biplanar-slide} pauses; the robot executes pivoting and flipping to avoid leaving a finger beneath the object, resumes \textsc{biplanar-slide} until the object settles in the slot, and finally readjusts the grasp to push the object to the target pose (Fig.~\ref{fig:fig7}h). The full procedure is repeated ten times on hardware and succeeds in all trials, demonstrating reproducibility.

\subsection{Generalization Ability of Our Framework}
\label{subsec:generalization}

We evaluated the framework’s generalization across diverse shapes and environments. For shape variation, the planner successfully generates valid manipulation plansfor 11-sided (Task 1) and 9-sided (Task 2) concave polygons. For environment variation, when the slot length in task~2 is modified, the planner still returns a feasible plan. Fig.~\ref{fg1} can be regarded as another case of environment variations of task 2, where the object is initially at the corner. These results indicate that the library–planner coupling adapts to geometric changes without altering the algorithmic pipeline.

We also evaluated the inverse of both tasks. In the inverse of task~1, the object is initially lifted by a zero-mobility gripper and the target is to place it on the surface; our framework produces an accurate plan that can be repeatedly executed on the real robot for ten trials. In the inverse of task~2, the object starts inside the slot and the goal is to grasp and suspend it in the air; the planner succeeds in eight out of ten trials. The two failures occurred during the biplanar-slide phase due to contact loss under our basic force/torque-gated controller; active force control would likely resolve this. Detailed executions and plans are available in the accompanying video.
\subsection{Necessity Validation of Manipulation Primitives}
\label{subsec:ablation}

To assess whether each manipulation primitive in our library is fundamentally required, we conducted ablation studies by removing a single primitive at a time and testing whether the two tasks in Sec.~\ref{sec:experiments} still admit feasible plans and successful executions. Table~\ref{tab:mp_ablation} summarizes the outcomes.

For \emph{task~1}, the planner produced feasible plans only when \textsc{slide} was removed; the resulting plans reached a $10/10$ success rate on hardware. Removing any other primitive led to no valid solution. For \emph{task~2} (slot insertion), feasible plans remained obtainable when \textsc{tip}, \textsc{close}, or \textsc{slide} was removed, and these plans also achieved a $10/10$ success rate; removing any other primitive caused failure. These results quantitatively confirm that most primitives are indispensable to complete the tasks, with the few exceptions reflecting redundancy that the planner can compensate for in specific geometries. 

The observations align with the functional roles of the primitives. Within \emph{MoveObject}, \textsc{tip} is essential for changing an object’s orientation while maintaining environment contact; \textsc{push} is necessary for in-contact translation; \textsc{biplanar-slide} adjusts orientation without leaving the two-edge contact state; and \textsc{air-move} transitions between contact and non-contact states. Within \emph{AdjustGrasp}, the set (\textsc{open}, \textsc{close}, \textsc{pivot}, \textsc{slide}, \textsc{flip}, \textsc{approach-contact}) covers the relative motions required for establishing, switching, and refining grasps. Notably in \emph{task~2}, without \textsc{flip} a finger becomes trapped beneath the object, blocking the biplanar-slide phase and preventing the object from being laid down. 


\section{CONCLUSIONS}
We introduced LDHP, a gripper-aware, library-driven planner that prioritizes executability in non-prehensile manipulation. The same pipeline executed on hardware across two distinct tasks (zero-mobility lifting and slot insertion) without re-design, adapted to variations in shape and environment dimensions, and ablations confirmed that most primitives are indispensable. These results suggest that gripper-aware primitive libraries, paired with layered search, offer a practical path to long-horizon, contact-rich manipulation that remains physically grounded and interpretable under modest sensing. Moving forward, we plan to extend LDHP to SE(3) settings, integrate force control for higher precision, and incorporate learned priors to accelerate search.

\begin{table}[t]
\centering
\caption{Ablation of manipulation primitives in two tasks (Y=feasible, N=not feasible; value in parentheses = successes /10).}
\label{tab:mp_ablation}
\scriptsize
\setlength{\tabcolsep}{10pt}      
\renewcommand{\arraystretch}{0.96} 
\begin{tabular}{lcc}
\toprule
\textbf{Primitive removed} & \textbf{Case 1} & \textbf{Case2} \\
\midrule
\multicolumn{3}{l}{\emph{MoveObject} family} \\
\midrule
\textsc{tip}               & N      & Y (10) \\
\textsc{push}              & N      & N      \\
\textsc{biplanar-slide}    & N      & N      \\
\textsc{air-move}          & N      & N      \\
\midrule
\multicolumn{3}{l}{\emph{AdjustGrasp} family} \\
\midrule
\textsc{open}              & N      & N      \\
\textsc{close}             & N      & Y (10) \\
\textsc{pivot}             & N      & N      \\
\textsc{slide}             & Y (10) & Y (10) \\
\textsc{flip}              & N      & N      \\
\textsc{approach-contact}  & N      & N      \\
\bottomrule
\end{tabular}
\vspace{-20pt}
\end{table}

\bibliographystyle{ieeetr}
\bibliography{references}

\end{document}